\documentclass[letterpaper]{article} 
\usepackage{aaai25}  
\usepackage{times}  
\usepackage{helvet}  
\usepackage{courier}  
\usepackage[hyphens]{url}  
\usepackage{graphicx} 
\usepackage{natbib}  
\usepackage{caption} 
\usepackage{algorithm}
\usepackage{algorithmic}
\usepackage{multirow}
\usepackage{booktabs}
\usepackage{amsmath}
\usepackage{amssymb}

\usepackage{newfloat}
\usepackage{listings}
\DeclareCaptionStyle{ruled}{labelfont=normalfont,labelsep=colon,strut=off} 
\floatstyle{ruled}
\newfloat{listing}{tb}{lst}{}
\floatname{listing}{Listing}
\title{ScreenMark: Watermarking Arbitrary Visual Content on Screen}
\author{
    Xiujian Liang\textsuperscript{\rm },
    Gaozhi Liu\textsuperscript{\rm },
    Yichao Si\textsuperscript{\rm }, 
    Xiaoxiao Hu\textsuperscript{\rm },
    Zhenxing Qian\textsuperscript{\rm}\thanks{Corresponding author.}
}
\affiliations{
    \textsuperscript{\rm }Fudan University, School of Computer Science\\
    ShangHai 200438, China\\

    liangxj23,gzliu22,ycsi22,xxhu23@m.fudan.edu.cn, zxqian@fudan.edu.cn
}

\usepackage{bibentry}

\begin{document}

\maketitle
\begin{abstract}
Digital watermarking has shown its effectiveness in protecting multimedia content. However, existing watermarking is predominantly tailored for specific media types, rendering them less effective for the protection of content displayed on computer screens, which is often multi-modal and dynamic. \textbf{Visual Screen Content (VSC)}, is particularly susceptible to theft and leakage through screenshots, a vulnerability that current watermarking methods fail to adequately address.
To address these challenges, we propose \textbf{ScreenMark}, a robust and practical watermarking method designed specifically for arbitrary VSC protection. 
ScreenMark utilizes a three-stage progressive watermarking framework. Initially, inspired by diffusion principles, we initialize the mutual transformation between regular watermark information and irregular watermark patterns. Subsequently, these patterns are integrated with screen content using a pre-multiplication alpha blending technique, supported by a pre-trained screen decoder for accurate watermark retrieval. 
The progressively complex distorter enhances the robustness of the watermark in real-world screenshot scenarios. 
Finally, the model undergoes fine-tuning guided by a joint-level distorter to ensure optimal performance.
To validate the effectiveness of ScreenMark, we compiled a dataset comprising 100,000 screenshots from various devices and resolutions. Extensive experiments on different datasets confirm the superior robustness, imperceptibility, and practical applicability of the method.
\end{abstract}

%

\section{Introduction}

With the continuous advancement of the Internet and computer technologies, an increasing amount of information is presented in the form of \textit{Visual Screen Content} (VSC), including images, videos, texts, web pages, windows, and more. 
In the context of users' personal computers, VSC is displayed on screens without exception. 
From a visual perspective, this mode of presentation is the most readily acceptable form of expression.
At the same time, this also means that the VSC can easily be leaked.

\begin{figure}[t]
  \includegraphics[width=\linewidth]{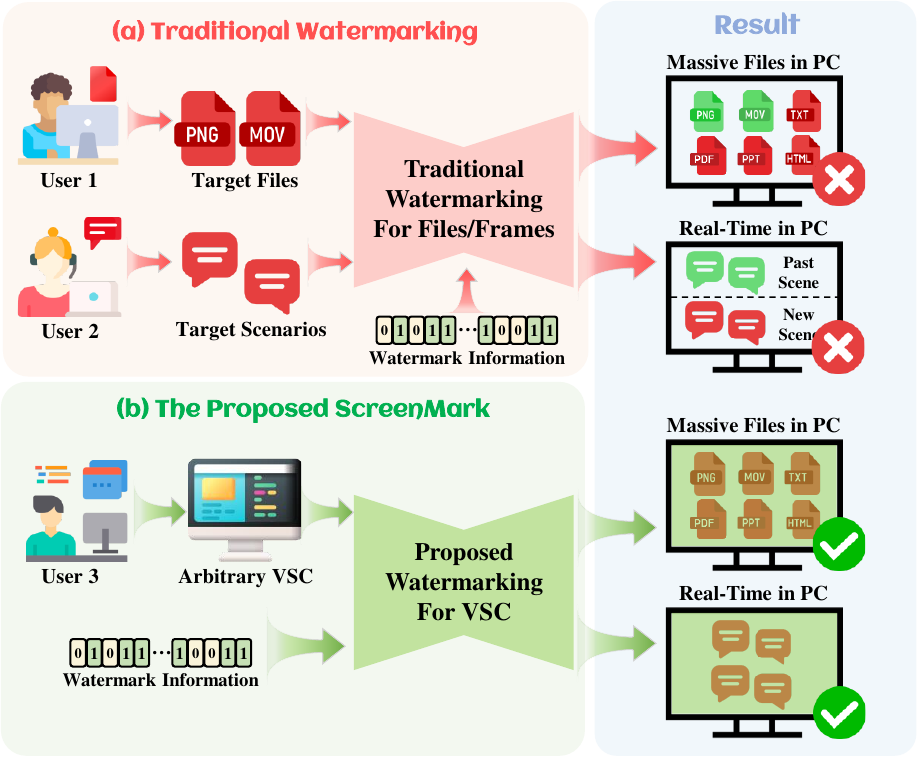}
  \caption{The comparison of the traditional watermarking and the proposed ScreenMark. The red and green content represent non-watermarked and watermarked, respectively. }
  \label{fig:teaser}
\end{figure}

For most enterprise and home computers, ensuring data security is a significant challenge. Traditional security measures, such as data encryption, firewalls, access control, and identity management, provide comprehensive protection against data leaks. However, these methods primarily manage permissions, allowing authorized users to capture screen content in real-time using screenshot tools.
Current specialized protection techniques for VSC often rely on non-learning watermarking methods, which are even visible. These methods struggle to balance robustness and visual quality effectively, rendering them unsuitable for real-world applications.
Therefore, this study focuses on VSC security in screenshot scenarios and aims to propose a universal learning-based screen watermarking method.

In recent years, the rapid advancement of multimedia watermarking technology\cite{liu2023wrap,xiao2024client,tang2024robust,li2024screen} has facilitated the protection of multimedia file content. 
However, it is important to recognize that current watermarking techniques are primarily designed for individual modalities, offering specialized protection for images, videos, text, and other media types.
As depicted in Fig.\ref{fig:teaser}, there are two main drawbacks: limited scope of protection and response time constraints.
Fig.1(a) shows that while traditional watermarking can safeguard specific media content within a single modality or frame, it falls short in providing comprehensive protection across various file types and the vast number of files present on personal computers. Moreover, these methods struggle to counteract millisecond-level capture attacks under dynamic screen conditions.
In contrast, the method proposed in this paper, illustrated in Fig.1(b), does not focus on protecting a single media file or specific screen frame at a given time. 
Instead, it integrates the watermark with the screen through a unique fusion process, offering comprehensive and real-time protection for arbitrary VSC displayed on the screen. This approach has been named \textbf{ScreenMark}.

In ScreenMark, we introduce a three-stage watermarking framework utilizing progressive training\cite{li2022automated}. This approach completes robustness training at various levels across different stages, resulting in a versatile screen watermarking solution for VSC sreenshot scenarios.
Traditional watermarking methods often embed watermark information directly into media content via an encoder, leading to increased processing time and limited protection scope.
To overcome these limitations, ScreenMark employs an irregular watermark pattern that blends more naturally and comprehensively with screen content, resembling a mask.
Moreover, this irregular pattern makes it more challenging for unauthorized users to detect and remove the watermark, thereby enhancing the security of the protection mechanism. 
Building on this, the three-stage progressive training strategy further refines this approach. Each stage addresses specific challenges: from basic message diffusion and reversal to adaptive screen decoder training, and finally, handling composite distortion.
Through progressively complex training scenarios, ScreenMark enhances system resilience in VSC capture and subsequent processing scenarios.

Based on the above, the contribution of this paper can be summarized as follows:

    1) We introduce a novel and practical multimedia protection scenario that addresses not only single-modal media content but also multi-modal VSC displayed on computer screens.  And we point out the limitation of the mainstream single-modal watermarking methods in terms of protection scope and response time in this scenario.
    
    2) To the best of our knowledge, we present the first learning-based watermarking framework specialized for VSC protection, named \textbf{ScreenMark}. In ScreenMark, regular watermarking information is diffused into irregular watermarking patterns and integrated with the screen display. We propose a three-stage progressive training strategy and design various levels of distorters tailored to different stage.
    
    3) To enhance the applicability of ScreenMark to VSC protection, we have compiled a dataset of 100,000 screenshot images. These images were collected using different screenshot tools across a diverse range of devices and resolutions from SD (720x480) to 4K (3840x2160).
    
    4) Extensive experiments demonstrate that ScreenMark matches or even surpasses the performance of four SOTA single-modal watermarking methods in screenshot scenarios in terms of robustness, invisibility, and applicability to real-world situations.

\section{Related Work}
\subsection{Deep-learning-based Watermarking}
Deep-learning approaches have effectively addressed the limitations of hand-crafted features in watermarking. 
\cite{zhu2018hidden} introduced an end-to-end solution using an auto-encoder architecture, establishing a foundation in the domain.  
To enhance robustness against JPEG compression, MBRS\cite{jia2021mbrs} proposed a hybrid noise layer of real and simulated JPEG with a small batch strategy. \cite{fernandez2022watermarking} utilized a pre-trained model to create a transform-invariant latent space for watermark embedding, achieving higher robustness against various attacks. DWSF\cite{guo2023practical} offers a practical framework for decentralized watermarking, training an auto-encoder to resist non-geometric attacks and incorporating a watermark synchronization module for geometric attacks.  Moreover, some recent methods\cite{guan2022deepmih,xu2022robust,fang2023flow} explore invertible neural networks for watermark embedding and extraction. However, these primarily protect specific media content in a single modality and are inadequate for the multi-modal VSC in real-time scenarios.

\subsection{Screen-related Watermarking}

The visibility of VSC to the public has sparked interest in its preservation among scholars. \cite{piec2014real} utilized the Human Visual System to create dynamically adaptable watermarks, while \cite{du2018adaptive} aimed to prevent screenshot data leakage through full-screen protection. These non-learning-based VSC watermarking methods struggled with balancing robustness and invisibility, making them easily detectable by attackers.

Driven by the need for screen protection, screen-shooting resilient watermarking (SSRW) addresses cross-channel content leakage. \cite{fang2018screen} first modeled screen shooting distortions and proposed robust watermarking schemes based on DCT and SIFT. \cite{wengrowski2019light} introduced CDTF, a network simulating the screen-to-camera process using a multi-million dataset. \cite{tancik2020stegastamp} developed a noise layer for the printer-to-camera channel, addressing various distortions. \cite{jia2020rihoop} designed a 3D reconstruction-based noise layer for the camera-shot channel, achieving camera-shot robustness. Similarly, \cite{fang2022pimog} created PIMoG, a noise layer for the screen-to-camera channel, enhancing screenshot robustness. To improve visual quality, \cite{jia2022learning} embedded information in sub-images and used a localization network to identify watermarked regions. To address distortion variability across different screens, \cite{fang2023denol} introduced DeNoL, an efficient decoupling noise layer that simulates distortions accurately with fewer samples by fine-tuning transform layer.

However, SSRW, which focuses on modeling the physical distortion of recapture, protects only specified fixed content and is limited in scope and response time, making it ineffective for screen interception scenarios in VSC. In contrast, our work aims to protect arbitrary VSCs, providing real-time protection as the screen changes and multi-modal protection within a single watermark framework.

\begin{figure*}[htbp]
  \includegraphics[width=\textwidth]{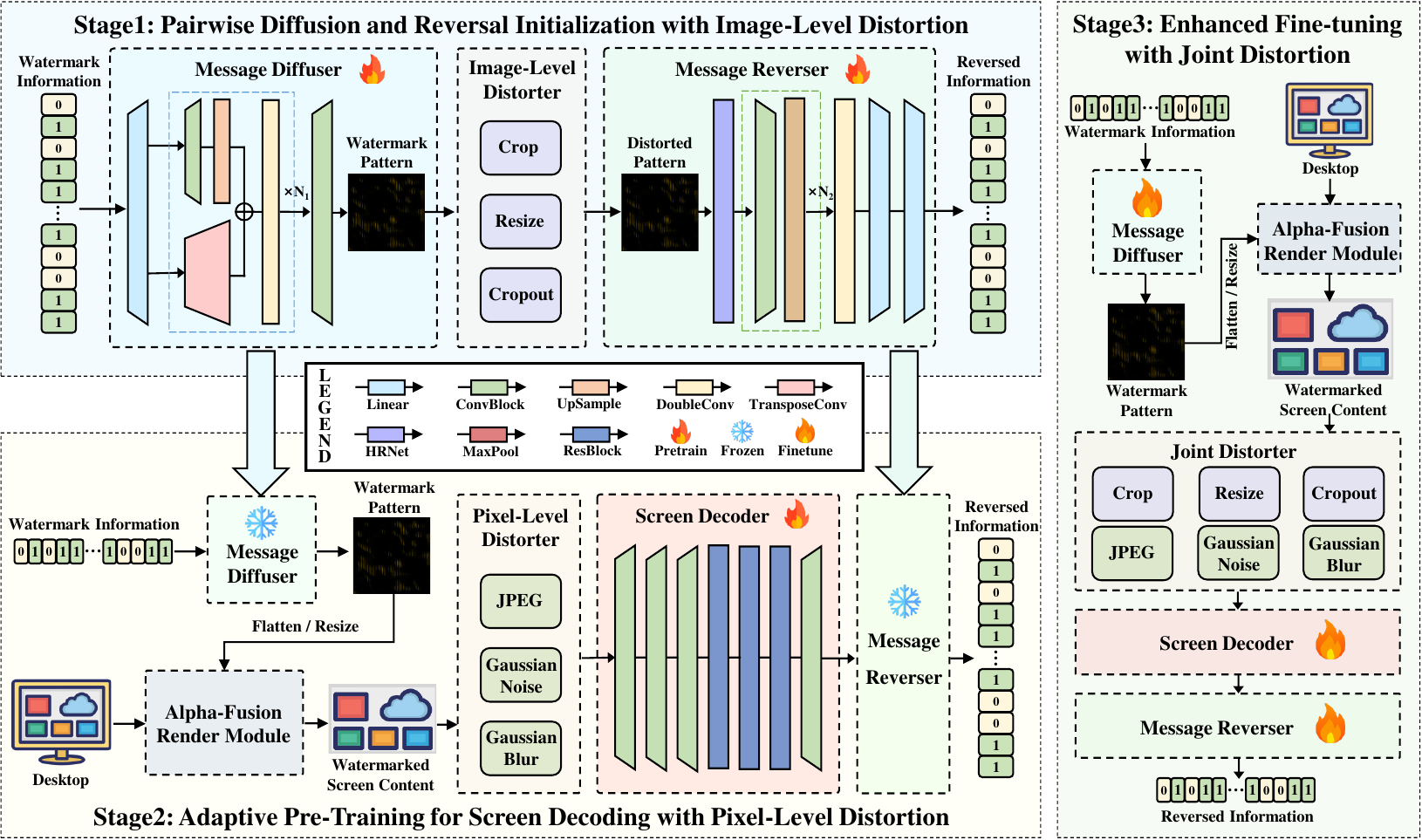}
  \caption{The overall framework of the proposed method, which contains three main stages.}
  \label{FIG2}
\end{figure*}

\section{Proposed Method}
\subsection{Motivation \& Overview}
To address the limitations in protection scope and response times encountered by current watermarking techniques for screen content, this paper introduces a three-stage watermarking framework specifically designed for screen content.
Inspired by the diffusion model\cite{ dhariwal2021diffusion}, which diffuses a regular image with noise to generate an irregular image, this framework adopts a novel approach. 
The diffusion process can be reversed through denoising, restoring the original image. This suggests directly integrating irregular watermark patterns obtained from the diffusion of regular watermark information with the screen content, rather than embedding information into the protected multimedia carrier using an encoder, as is common in traditional watermarking frameworks.
This is particularly important for screen content protection because it allows real-time and cost-effective integration of watermark patterns with screen content. It offers a protection method that is unrestricted by scope and response time, integrating securely and naturally with screen content.

Our approach transforms regular watermark information into irregular watermark patterns and integrates them with screen content, eliminating the need for encoder-based information embedding. 
With this in mind, we designed a three-stage watermarking framework, executing robustness training at different levels. The overall framework shows in Fig.\ref{FIG2}.

\subsection{Stage-1: Pairwise Initialization}
In \textbf{Stage-1}, we initialize a pair of a \textit{Message Diffuser} and a \textit{Message Reverser}. These modules effectively facilitate the dissemination of watermark information into watermark patterns and their subsequent reverse recovery. During this stage, we introduce an \textit{image-level distorter} to enhance robustness against image-level attacks that may occur with screen captures.The framework consists of a Message Diffuser ${M_D}$, Message Reverser ${M_R}$, and an Image-level Distorter ${D_I}$.

\subsubsection{Workflow}
Initially, we generate a batch of regular watermark messages ${I_w}$ , with batch size ${N_0}$ and length of information ${L}$. Subsequently, ${I_w}$ is subjected to diffusion processing within ${M_D}$, resulting in an irregular watermark pattern ${P_w}$. Upon acquisition of ${P_w}$, each pattern in batch undergoes parallel image-level distortion processing by ${\mathcal{D}_I^k \in D_I}$, where ${k \in \{1, 2, \ldots, N_0\}}$, yielding the distorted watermark pattern ${P_d}$. Finally, the reverse-processed watermark information ${I_r}$ is recovered through ${M_R}$.

\subsubsection{Architecture}
The ${M_D}$ is comprised of a linear layer, ${N_1}$ diffusion block, and a convolution Block connected in series. It receives a regular watermark information ${I_w}$ of size ${\mathbb{R}^{ 1\times 1\times L\times N_0}}$ and outputs a regular watermark pattern ${P_w}$ of size ${\mathbb{R}^{ H\times W\times 3\times N_0}}$. 
The diffusion blocks use upsampling and transposed convolution in parallel to suppress checkerboard effects and enhance feature representation without increasing network depth.
Subsequently, in ${D_I}$, different batches of ${I_w}$ are randomly subjected to one of three types of image-level distortions ${\mathcal{D}_I^k \in D_I}$, specifically Resize distortion, Crop distortion , and Cropout distortion. 
These distortions are common in actual screenshot scenarios and cannot be restored by third parties without the attack parameters. 
Conversely, the  ${M_R}$ consists of an HRNet,  ${N_2}$ reversal block, a double-convolution block, and two linear layers connected in series. It receives a distorted watermark pattern ${P_d}$ of size ${\mathbb{R}^{ H\times W\times 3\times N_0}}$ and outputs a reversed watermark information ${I_r}$ of size ${\mathbb{R}^{ 1\times 1\times L\times N_0}}$.

\subsubsection{Loss Function}
In this stage, the loss function serves two purposes: on one hand, control the security and stealthiness of the generated watermark patterns, and on the other hand, ensure the accuracy of the reversed information. To achieve the irregularity and invisibility of the watermark patterns, we propose four types of losses: near-zero loss, dispersion loss, variation loss, and channel loss.

\textbf{Near-zero loss} reduce the interference of the watermark patterns on the original screen content. It minimize the mean squared error between the generated watermark pattern ${P_d}$ and a tensor filled with zeros, signifying that the overall pixel values are close to zero , which can be
formulated as:
\begin{equation}
 L_{zero} = MSE(P_d, \mathbf{0})
\label{L_Zero}
\end{equation}
where \textbf{0} represents the zero matrix with shape as ${P_d}$. 

\textbf{Dispersion loss} prevent concentration in specific areas for robustness against image processing attacks. It increases the dispersity of the watermark pattern by calculating the mean absolute value and variance, ensuring uniform distribution across the image, which can be
formulated as:
\begin{equation}
L_{dispersion} = mean(|P_d|) + var(P_d)
\label{L_Dispersion}
\end{equation}
where ${|P_d|}$, ${mean}$ and ${var}$ represents the absolute value, the mathematical mean and the variance of ${P_d}$, respectively. 

\textbf{Variation loss} reduce the visual noise introduced by the watermark. It encourages the spatial smoothness of the generated watermark pattern by minimizing drastic changes between adjacent pixels, making the watermark harder to detect by the naked eye, which can be formulated as:
\begin{equation}
\begin{split}
L_{variation}=& \sum_{i,j}^{W,H}[\sqrt{(P_d[i, j] - P_d[i+1, j])^2}+ \\ 
              & \sqrt{(P_d[i, j] - P_d[i, j+1])^2}]
\label{L_Variation}
\end{split}
\end{equation}
where $i$ and $j$ respectively represent the row and column indices of pixels in the ${P_d}$. $W$ and $H$ represent the width and height of the ${P_d}$. Variation loss considers pixel variations in both the vertical and horizontal directions, reducing high-frequency noise while preserving edge information.

\textbf{Channel balance loss} reduce noticeable color distortion and maintain the color balance. It minimizes the mean squared error of the mean value interpolation of the R, G, B color channels. This design is particularly important for color-sensitive screen content, which can be formulated as:
\begin{equation}
\begin{split}
L_{Channel} = &  mean((R_{m} - G_{m})^2 + \\
              &  (R_{m} - B_{m})^2 + (G_{m} - B_{m})^2)
\label{L_channel}
\end{split}
\end{equation}
where ${R_{m}}$,${G_{m}}$ and ${B_{m}}$ respectively represent the mean of R,G and B channel of ${P_d}$, respectively.

Based on the considerations above, the corresponding loss function for the pattern can be written as follows:
\begin{equation}
\begin{split}
L_{Pattern} = & \lambda_{0}L_{zero} + \lambda_{1}L_{dispersion} + \\
              & \lambda_{2}L_{variation} + \lambda_{3}L_{channel}
\label{L_Pattern}
\end{split}
\end{equation}

To ensure the consistency of reversed information with the watermark, we design the loss function for the message using binary cross-entropy, which can be formulated as:
\begin{equation}
\begin{split}
L_{Message} = & -\frac{1}{N}\sum_{i=1}^{N} [I_{w_i} \log(I_{d_i}) + \\
              & (1 - I_{w_i}) \log(1 - I_{d_i})]
\label{L_Watermark}
\end{split}
\end{equation}
where ${N}$ is the number of samples in the batch,  ${I_{d_i}}$ is the predicted value for the ${i^{th}}$ sample, and ${I_{w_i}}$ is the actual value for the ${i^{th}}$ sample. The loss function in the initialization training process of stage-1 can be formulated as:
\begin{equation}
L_{stage1} =  \beta L_{Pattern} + \gamma L_{Message}
\label{L_stage1}
\end{equation}
where $\beta$ and $\gamma$ each represent the importance of watermarking patterns and watermarking message in this stage.
\begin{table*}[thbp]
\centering
\renewcommand\arraystretch{1.0}
\resizebox{\linewidth}{!}{
\begin{tabular}{cc|c|c|ccc|ccc|ccc|c} 
\toprule[1.5pt]
\multirow{2}{*}{Dataset}  & \multirow{2}{*}{Methods}  & \multirow{2}{*}{\begin{minipage}{1cm} \centering Length \\ (Bits) \end{minipage}} & \multirow{2}{*}{Clean} & \multicolumn{3}{c|}{Crop(\%)} & \multicolumn{3}{c|}{Cropout(\%)} & \multicolumn{3}{c|}{Resize(\%)} &  \multirow{2}{*}{Average} \\
\cmidrule{5-13}
                          & & &                                     & 90 & 80 & 70             & 5 & 10 & 20        & 80 & 150 & 300      &  \\ 
\midrule[1.0pt]
\multirow{5}{*}{ImageNet} & StegaStamp   &\textbf{100}   &\underline{99.48}     &68.93  &55.19  &51.93     &97.12  &97.00 &\underline{95.31}    &91.00  &91.43    &90.12      &82.00  \\
                            & MBRS      &30    &97.97               &79.64     &75.43   &72.13    &97.36    &97.11  &93.87      &93.38   &93.16     &92.64      &88.30   \\
                          & PIMoG     &30  &99.38                   &78.83  &73.14   &70.16      &97.56    &97.31  &93.12     &92.21    &92.17     & 90.25    &87.19  \\
                          & DWSF      &30   &\textbf{100.00}       &\textbf{98.75}   &\underline{96.58}   &\underline{94.12}     &\textbf{98.16}    &\textbf{97.89}  &94.44      &\underline{94.35}    &\underline{97.37}    &\textbf{97.73}      &\underline{96.60}   \\
                          & ScreenMark   &\textbf{100 } &\textbf{100.00}  &\underline{98.71} &\textbf{96.75} &\textbf{94.17}      &\underline{98.09}  &\underline{97.38}  &\textbf{95.65}     &\textbf{95.30}   &\textbf{97.98}    &\underline{97.25}    & \textbf{96.80}             \\ 
\midrule[1.0pt]
\multirow{5}{*}{ScreenImage} & StegaStamp  &\textbf{100}    &97.06   &72.68   &60.25  &55.93       &90.43  &91.06  &89.00         &94.68  &93.68    &94.31   &82.45     \\
                          & MBRS        &30     &98.12   &79.99 &75.80 &72.46 &97.79 &97.54 &94.22 &93.73 &93.51 &92.99 &88.67             \\
                          & PIMoG       &30      &\underline{99.45}   &79.19 &73.40 &70.45 &97.99 &97.75 &93.47 &92.56 &92.52 &90.59 &87.55             \\
                          & DWSF        &30    &\textbf{100.00}      &\underline{99.20} &\underline{96.82} &\underline{93.46}        &\underline{98.61} &\underline{98.34} &\underline{95.08} &\underline{94.79} &\underline{95.82} &\underline{95.17} &\underline{96.37}               \\
                          & ScreenMark   &\textbf{100}    &\textbf{100.00}      &\textbf{99.66} &\textbf{98.73} &\textbf{95.26}        &\textbf{99.93}  &\textbf{99.72}  &\textbf{96.23}      &\textbf{99.88}  &\textbf{99.99}  &\textbf{99.99}      &\textbf{98.82}         \\
\bottomrule[1.5pt]
\end{tabular}
}
\caption{Bit accuracy rate(BAR,\%) on different image-level attacks}
\label{tab:robust_img}
\end{table*}
\subsection{Stage-2: Adaptive Pre-Training}
In \textbf{Stage-2}, we involve adaptive training for the \textit{screen decoder} ${D_S}$ to accurately decode the irregular watermark patterns and reverse the message. We freeze the parameters of the model from stage 1 and input the patterns into an \textit{Alpha blending rendering module} ${R_{\alpha}}$ for integration with the computer screen. This stage also introduces a \textit{pixel-level distorter} to increase robustness against pixel-level attacks.

\subsubsection{Workflow}
The process of transforming watermark information into watermark patterns remains consistent with Stage-1. Building on this, at this stage, the watermark patterns $P_w$ and the screen content $S_c$ are input into ${R_{\alpha}}$. Through flattening or scaling, ${R_{\alpha}}$ integrates $P_w$ with $S_c$ to produce the watermarked screen content $S_w$. The specific way of integration is to form a kind of mask that floats above the screen through the $\alpha$ channel, so that the full-screen watermark effect can be realized on any resolution screen without any impact on the screen content. 
After obtaining $S_w$, we employ Pixel-level Distorter $\mathcal{D}_P^k \in D_P$ to process with parallel distortion, where ${k \in \{1, 2, \ldots, N_0\}}$, resulting in the distorted watermarked screen content $S_d$. Finally, the distorted watermark pattern $P_d$ is decoded from $S_d$ using ${D_S}$.
\subsubsection{Architecture}
The ${R_{\alpha}}$ utilizes Direct3D for efficient graphics processing and the Windows API for window management, incorporating pre-multiplied alpha technology to optimize transparency handling. Initially, window creation and configuration are performed using Windows' CreateWindowEx, ensuring the window stays atop others while allowing mouse events to pass through, maintaining unobtrusive user interaction. Subsequently, Direct3D is utilized for GPU-accelerated rendering, enhancing efficiency and ensuring speed during terminal information switching. Finally, Pre-multiplied alpha image processing is applied before loading, where each pixel's color value is multiplied by its alpha value, simplifying transparency blending calculations. This computation can be described as follows:
\begin{equation}
S_w = \alpha P_{w} + (255 - \alpha) S_c 
\label{Alpha}
\end{equation}
where ${\alpha}$ take the value of 5, meaning the watermark pattern affects less than 2\% of screen content pixels. To optimize the image rendering, an advanced shader dynamically adjusts image transparency during rendering, leveraging pre-multiplied alpha techniques for automatic color and transparency blending.
In ${D_P}$, batches of ${S_w}$ are randomly selected among three types of image-level distortions, denoted as $\mathcal{D}_P^k \in D_P$. These types include JPEG compression, Gaussian noise, and Gaussian blur.
The ${D_S}$ consists of three consecutive ConvBlock, two ResBlock, and an additional ConvBlock, linked in series. It receives ${S_d}$ of size ${\mathbb{R}^{ H\times W\times 3\times N_0}}$ and outputs a ${P_{ds}}$ of the same size.
\subsubsection{Loss Function}
This stage facilitates the ${D_S}$ in pretraining adaptively, building on the training foundation laid in the previous stage, to decode watermark patterns from screenshots of watermarked screens that have undergone pixel-level distortion attacks. The goal is to match the extracted pattern as closely as possible to the original pattern before distortion.  With the weights from Stage-1 frozen, pre-training+
for the screen decoder is not interfered by other factors. The loss function in stage-2 is formalized as follows:
\begin{equation}
L_{stage2} = MSE(P_{ds}, P_w)
\label{L_stage2} 
\end{equation}

\subsection{Stage-3: Enhancement Fine-Tuning}
In \textbf{Stage-3}, we will synergistically fine-tune the model weights acquired from the Message Diffuser ${M_D}$, Message Reverser ${M_R}$, and Screen Decoder ${D_S}$ from the previous two stages. Furthermore, we introduce an additional Joint-level Distortion Layer ${D_J}$ that encompasses both image and pixel-level distortions. This enhances the model's robustness when integrated with screen content, effectively compensating for the limitations of the previous two stages.
In order to make the model have just the right amount of robustness, the level of the distorter and its position are different.
Based on the aforementioned model, we have named the complete network \textbf{ScreenMark}. The loss function during the enhancement fine-tuning process is as follows:
\begin{equation}
L_{stage3} = L_{stage1} + L_{stage2}
\label{L_stage3} 
\end{equation}

\section{Experiments}
\subsection{Experimental Settings}
\subsubsection{Benchmarks.} We are the first learning-based watermarking specialized for VSC protection and have no directly relevant baseline model to compare against. In order to measure our performance in terms of robustness, we still compared our method with four state-of-the-art(SOTAs) single-modal watermarking methods, i.e.,Stegastamp\cite{tancik2020stegastamp}, PIMoG\cite{fang2022pimog}, MBRS\cite{jia2021mbrs}, DWSF\cite{guo2023practical}.

\begin{table*}[thbp]
\centering
\renewcommand\arraystretch{1.0}
\resizebox{\linewidth}{!}{
\begin{tabular}{cc|c|c|ccc|ccc|ccc|c} 
\toprule[1.5pt]
\multirow{2}{*}{Dataset}  &\multirow{2}{*}{Methods}  &\multirow{2}{*}{\begin{minipage}{1cm} \centering Length \\ (Bits) \end{minipage}} & \multirow{2}{*}{Clean}  & \multicolumn{3}{c|}{JPEG(QF)} & \multicolumn{3}{c|}{Noise($\sigma$)} & \multicolumn{3}{c|}{Blur($\kappa$)} & \multirow{2}{*}{Average} \\
\cmidrule{5-13}
                  &&         &                          & 95 & 85 & 75              &0.01  &0.05 & 0.1          &3     &5     &7         &     \\ 
\midrule[1.0pt]
\multirow{5}{*}{ImageNet} & StegaStamp    &\textbf{100}    &98.48    &95.43   &94.18 &92.37     &97.31    &96.75   &91.62    &96.62    &95.37  &94.12   &94.86       \\
                          & MBRS       &30    &97.97       &97.45     &96.53    &95.47    &98.29   &98.17   &\textbf{97.88}     &97.43   &95.89   &94.14       &96.80     \\
                          & PIMoG       &30    &\underline{99.38}       &96.38   &95.14  &93.27    &98.38    &98.29   &\underline{97.64}    &97.75    &\textbf{96.43}    &95.13  &96.49     \\
                          & DWSF      &30    &\textbf{100.00}       &\textbf{99.31}     &\underline{97.64}    &\textbf{95.88}    &\textbf{99.13}  &\underline{98.65}   &95.24      &\underline{98.66}  &96.17    &\textbf{96.82}       &\textbf{97.50}    \\
                          & ScreenMark      & \textbf{100} &\textbf{100.00}         &\underline{99.21} &\textbf{97.68} &\underline{95.77}   &\underline{99.03} &\textbf{98.68} &95.23        &\textbf{98.85}  &\underline{96.23}   &\underline{95.84}    & \underline{97.39}       \\ 
\midrule[1.0pt]
\multirow{5}{*}{ScreenImage} & StegaStamp   &\textbf{100}  &97.06     &93.00   &92.31  &90.31        &95.00   &89.12  &80.43      &93.68   &94.06    &93.81  & 91.30   \\
                          & MBRS     &30    &98.35 &98.92 &\underline{97.49} &\underline{95.58} &98.63 &98.51 &\textbf{98.18} &97.91 &96.35 &94.67       &\underline{97.36}        \\
                          & PIMoG     &30   &\underline{99.86} &96.89 &95.59 &93.47 &98.79 &\underline{98.70} &\underline{98.10} &98.22 &\underline{96.98} &95.59       &96.92       \\
                          & DWSF     &30       &\textbf{100.00}           &\underline{99.15} &97.38 &95.27     &\underline{99.21} &98.43 &95.12      &\underline{99.45} &96.20 &\textbf{95.86}    &97.34     \\
                          & ScreenMark      & \textbf{100} &\textbf{100.00}         &\textbf{99.39} &\textbf{98.34} &\textbf{96.69}        &\textbf{100.00}  &\textbf{99.08}  &96.19     &\textbf{100.00}   &\textbf{98.38}    &\underline{95.65}   &\textbf{98.19}        \\
\bottomrule[1.5pt]
\end{tabular}
}
\caption{Bit accuracy rate(BAR,\%) on different pixel-level attacks}
\label{tab:robust_pixel}
\end{table*}

\subsubsection{Datasets.}Given the absence of a suitable screenshot dataset for VSC protection, we created a dataset called ScreenImage, comprising 100,000 screenshots from various devices and resolutions ranging from SD (720x480) to 4K (3840x2160). We randomly selected 50,000 images as our training dataset. To evaluate the ScreenMark, we randomly sample 1,000 images each from ImageNet\cite{deng2009imagenet} and ScreenImage(excluding training) respectively. Notably, MBRS only accepts a fixed input size post-training, necessitating the scaling of test images to 128x128. Detailed dataset categorization and collection are displayed in APPENDIX.
\subsubsection{Implementations.} Our method is implemented using PyTorch\shortcite{paszke2019pytorch} and executed on an NVIDIA GeForce RTX 4090 GPU.  In terms of experimental parameter, the information length ${L}$ is 100. The height ${H}$ and width ${W}$ of watermark pattern $P_W$ is 512, optimized for blending with screens of different resolutions without implying a minimum size limit. The batch size ${N_0}$, diffusion block ${N_1}$ and reversal block ${N_2}$ is 16, 5 and 2, respectively. In Stage-1, the loss function weight factors $\beta$ and $\gamma$ are 0.1 and 1, respectively. For $L_{Pattern}$ , $\lambda_{0}$, $\lambda_{1}$, $\lambda_{2}$ and $\lambda_{3}$ are set to 1.0, 0.5, 0.1 and 0.01, respectively. Through 8 sets of ablation studies, we balanced choices based on fast training and high visual quality. The $\alpha $ of Alpha-Fusion Rendering Module ${R_{\alpha}}$ is set to 5, allowing us to control the range by adjusting $\alpha $ to achieve a trade-off. Robustness remains stable when the PSNR is between 36 and 42 dB ($\alpha$$\in[5,8]$), which aligns with the PSNR range of the SOTAs. We use the Adam optimizer\shortcite{kingma2014adam} with a learning rate of 1e-5, and set the training epochs to 100, while the compared methods adopt their default settings. Hyperparameter ablations are provided in APPENDIX.

\subsubsection{Metrics.} We consider Peak Signal to Noise Ratio (PSNR), Structure Similarity Index Measure (SSIM), Learned Perceptual Image Patch Similarity\cite{zhang2018unreasonable} (LPIPS) to evaluate visual quality. And consider Bit Accuracy Rate (BAR) evaluate robustness performance.

\subsection{Robustness Performance}
In this section, we compare the robustness of our ScreenMark method with four SOTAs across various attack types. The watermark length was set to 100 bits in our experiments. The types and implementation details of the attack settings also align with those used in SOTAs. To ensure objective results, we used the ImageNet and ScreenImage datasets for our experiments. Further experiments on watermark length, robustness against severe, hybrid and real-world scenarios attacks are included in the APPENDIX.

\subsubsection{Robustness against Image-level Attacks} 
We evaluate the robustness of the ScreenMark and SOTAs against image-level attacks in different factors. In Tab.\ref{tab:robust_img}, the headers represent the proportion of the Crop, Cropout, and Resize attacks in the original images, measured in percentage. The attack methods used in experiments align with the SOTAs for consistency and comparability. Notably, Our ScreenMark consistently achieves over 94\% bit accuracy, with an average performance of  97.81\% across both datasets, regardless of the attack type. Although in some cases do not outperform the best method, our performance remains close to the top.

\subsubsection{Robustness against Pixel-level Attacks} 
In addition to image-level attacks, we also assessed robustness against pixel-level attacks, which are common in social networking scenarios. Tab.\ref{tab:robust_pixel} reports the bit error rate against pixel-level attacks in different factors.  The table headers indicate the factors for each attack: JPEG compression quality factor (QF), Gaussian noise standard deviation ($\sigma$), and Gaussian blur kernel size ($\kappa$). Our ScreenMark method demonstrates exceptional stability and performance across various pixel-level attacks. In many cases, ScreenMark achieves the highest or second-highest bit accuracy rates, highlighting its robustness. Overall, ScreenMark consistently performs at a level comparable to the best method, and often surpasses other methods by approximately 1\%, showcasing its reliability and effectiveness.

\subsubsection{Robustness in Real screenshot Scenarios} 
To verify the practical application of ScreenMark, we tested its robustness in real screenshot scenarios. We collected screenshots using various publicly available tools, including Windows Screenshot, Snipaste, Greenshot, and WeChat Screenshot, across different resolutions. In experiments, watermarked VSC screenshots were randomly cropped with a fixed cropping size of 400*400 pixels, not exceeding 8\% of the area of a 1080P image. The screenshots were saved in JPG format, with the compression quality determined by the default settings of each tool. The results are inclued in APPENDIX, indicating that ScreenMark maintains a bit accuracy rate above 94\% across all resolutions and tools. This level of accuracy ensures that the watermark can be fully and correctly extracted in practical applications, especially when error-correcting codes are incorporated.


\subsection{Visual Quality}
The present work not only addresses the shortcomings of mainstream watermarking methods in VSC protection scenarios, demonstrating strong and stable robustness, but also achieves impressive visual quality. We verify the excellent performance of our work in terms of visual quality through qualitative visualization and quantitative metrics.
\begin{figure}[h]
\centering
  \includegraphics[width=\linewidth]{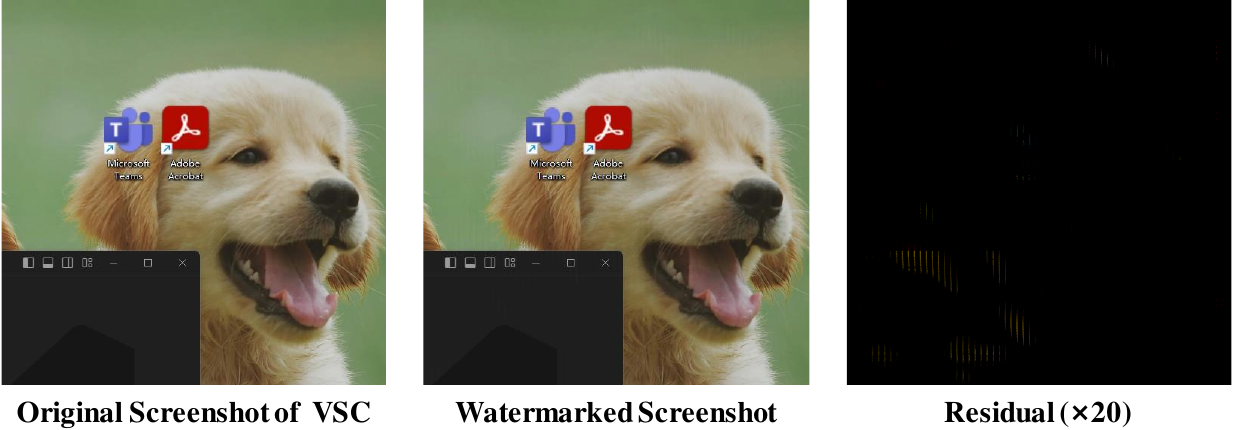}
  \caption{The visualization of the ScreenMark impact}
  \label{FIG3}
\end{figure}
\subsubsection{Visualization of Watermarking Residuals}
As described in Stage-1, we control the generated watermark pattern to be an irregular image that is close to zero, evenly dispersed, gently varying, and as balanced in RGB channels as possible. 
This minimizes the impact on the original VSC quality while avoiding malicious recognition and erasure by watermark attackers. 
Our proposed ScreenMark is fused with arbitrary VSC, protecting massive media contents on screen in real time that SOTAs cannot. 
Considering this, we calculated and magnified the residuals of the watermarked image compared to the original image by 20 times, using a randomly selected test image from ScreenImage. The results of the visualization are shown in Fig.\ref{FIG3}, confirming that our watermark pattern meets our design expectations. Richer visualization of ScreenMark is shown in APPENDIX.

\subsubsection{Quantification of Watermarked Images}
We introduced relevant image quality metrics from existing watermarking frameworks to assess visual quality. PSNR measures the image quality reference value between the maximum signal and background noise in dB, with higher values indicating less distortion. SSIM quantifies structural similarity between two images, with values from 0 to 1, and higher values indicating more similarity. LPIPS measures the difference between two images, with lower values indicating greater similarity. Table \ref{tab:visual_quantificationl} reports the visual quantification of watermarked images using different methods. ScreenMark achieves the best performance in both SSIM and LPIPS metrics, thanks to our pattern control and alpha fusion strategy.
\begin{table}[bthp]
\centering
\renewcommand\arraystretch{0.85}
\resizebox{0.95\linewidth}{!}{
\begin{tabular}{c|c|ccc} 
\toprule[1.5pt]
{Datasets}                 & {Methods}               &{PSNR$\uparrow$}    &{SSIM$\uparrow$}   &{LPIPS$\downarrow$}   \\
\midrule[1.0pt]
\multirow{5}{*}{ImageNet} & StegaStamp               &23.89           &0.8025        &0.0515    \\
                          & MBRS                     &36.49              &0.9173   &0.0387          \\
                          & PIMoG                    &36.21            &\underline{0.9850}        &0.0312          \\
                          & DWSF                     &\textbf{41.47}             &0.9831   &\underline{0.0083}          \\
                          & ScreenMark               &\underline{41.38}             &\textbf{0.9969}       &\textbf{0.0058}          \\ 
\midrule[1.0pt]
\multirow{5}{*}{ScreenImage} & StegaStamp               &28.06              &0.9411     &0.1640  \\
                          & MBRS                     &38.21              &0.9384         &0.0232         \\
                          & PIMoG                    &38.33            &0.9873         &0.0218         \\
                          & DWSF                     &\textbf{42.60}            &\underline{0.9905}         &\underline{0.0074}         \\
                          & ScreenMark               &\underline{41.86}             &\textbf{0.9948}   &\textbf{0.0055}         \\
\bottomrule[1.5pt]
\end{tabular}
}
\caption{Visual quantification in different methods}
\label{tab:visual_quantificationl}
\end{table}

\subsection{Other Comparison}
We have demonstrated the visual quality and robustness performance of ScreenMark, the key metrics for mainstream watermarkings. However, due to the unique scenarios addressed in this work, additional differences between ScreenMark and SOTAs need to be experimentally verified.

The first key difference is the advantage of the proposed three-stage progressive training strategy in ScreenMark compared to the traditional end-to-end training approach. The second difference lies in the scope of protection, where ScreenMark offers a broader range of coverage than single-modality watermarking, as depicted in Fig.\ref{fig:teaser}, so it will not be verified again. The third difference is the response time for watermark embedding in dynamic VSC changes, where ScreenMark significantly outperforms single-modality watermarking.

\subsubsection{The Ablation of Training Strategy}
\begin{figure}[t]
\centering
  \includegraphics[width=\columnwidth]{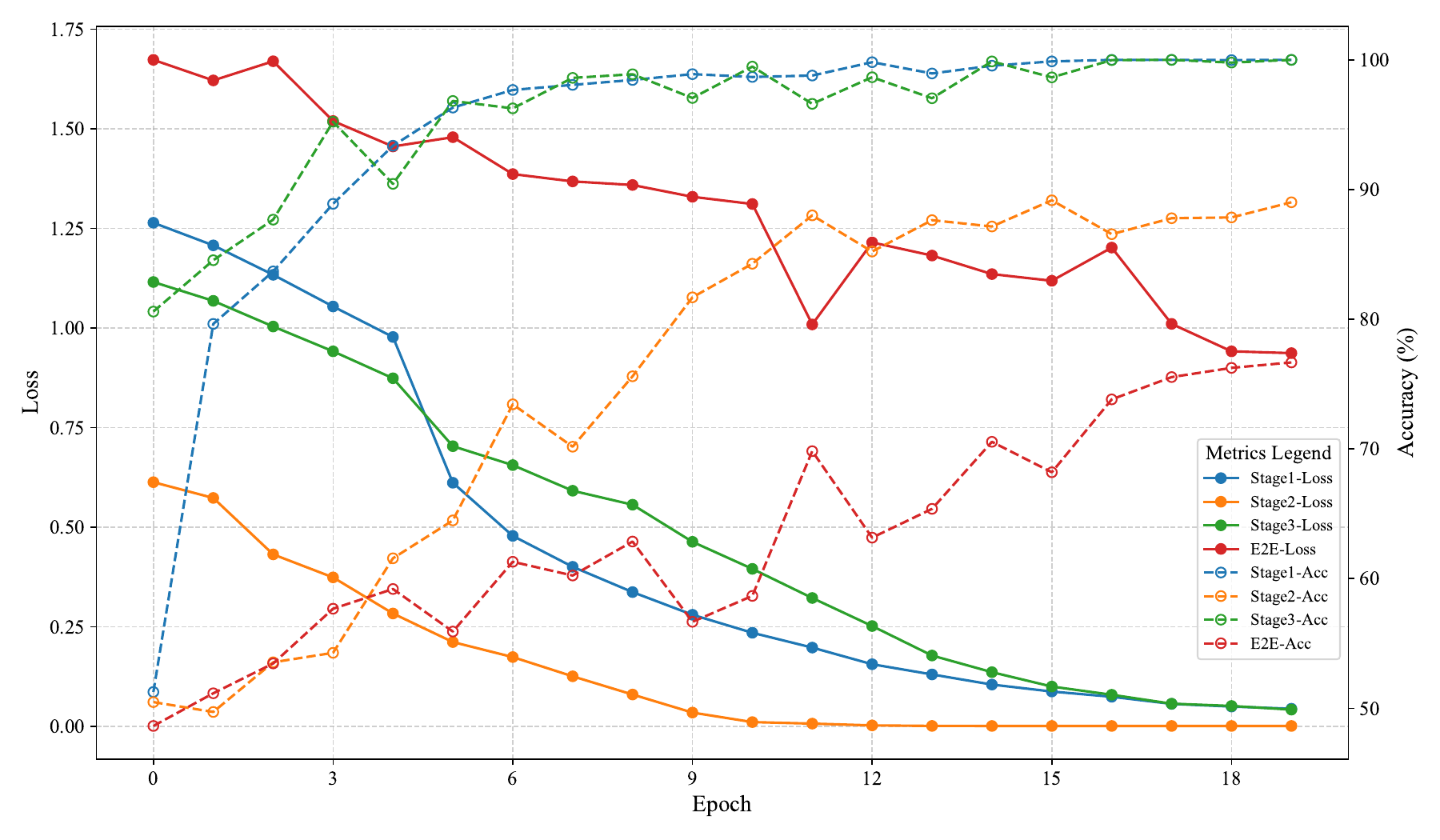}
  \caption{The Loss \& Acc vs. Epoch of training strategy}
  \label{FIG4}
\end{figure}

One key advantage of ScreenMark is its three-stage progressive training strategy, compared to the traditional end-to-end training approach. This strategy provides two main benefits: it allows the model to gain experience with simpler tasks first, simplifying the learning process for more complex tasks and preventing premature convergence to local optima. Additionally, staged training enables the use of more focused and refined loss functions and optimization strategies at each stage. We conducted an ablation study to compare different training strategies, as shown in Figure \ref{FIG4}. Our results indicate that the three-stage strategy achieves convergence by the 15th epoch, while E2E training exhibits oscillations in the loss function.

\subsubsection{The Comparison of Temporal Limitation}
Digital watermarking for VSC must be capable of real-time protection, as screen capture commands can be scripted to achieve \textbf{ millisecond level interception}. Existing single-modal watermarks can not embed in dynamic screen content in real time. We validated the responsiveness of SOTAs to VSC changes. As shown in Tab.\ref{tab:comparison_time}, our watermark, fused directly to the screen, achieves a reaction time of 0 milliseconds, outperforming SOTAs. In particular, here the time is not the execution time of program, but \textbf{the reaction time} to reload the watermark information to the new VSC when it changes.

\begin{table}[thbp]
\centering

\renewcommand\arraystretch{1.3}
\resizebox{\linewidth}{!}{
\begin{tabular}{c|ccccc} 
\toprule[1.5pt]
SOTAs      & StegaStamp   & MBRS &PIMoG   & DWSF & ScreenMark\\
\midrule[1.0pt]
Time(ms)   &24.91         &18.86  &19.83    &14.91  & \textbf{0}\\
\bottomrule[1.5pt]
\end{tabular}
}
\caption{Average reaction time of SOTAs in VSC scenarios}
\label{tab:comparison_time}
\end{table}

\section{Conclusion}
In this paper, we propose \textbf{ScreenMark}, a robust deep learning-based robust watermarking scheme for arbitrary VSC protection, using a three-stage progressive watermarking training strategy. 
The message diffuser and message reverser facilitate the transformation between regular watermarking information and irregular watermarking patterns. The alpha-fusion rendering module integrates these patterns into VSCs of any resolution, while the screen decoder extracts the watermark information from distorted watermarked screenshots. 
We built a dataset with 100,000 screenshots from various devices and resolutions. Extensive experiments demonstrate the effectiveness of ScreenMark in robustness, imperceptibility, and practical applicability.

\section{Acknowledgments}
This work was supported by the National Natural Science Foundation of China under Grants U20B2051, U22B2047, 62450067, 62072114.

\bibliography{aaai25}

\begin{thebibliography}{26}
\providecommand{\natexlab}[1]{#1}

\bibitem[{Deng et~al.(2009)Deng, Dong, Socher, Li, Li, and Fei-Fei}]{deng2009imagenet}
Deng, J.; Dong, W.; Socher, R.; Li, L.-J.; Li, K.; and Fei-Fei, L. 2009.
\newblock Imagenet: A large-scale hierarchical image database.
\newblock In \emph{2009 IEEE conference on computer vision and pattern recognition}, 248--255. Ieee.

\bibitem[{Dhariwal and Nichol(2021)}]{dhariwal2021diffusion}
Dhariwal, P.; and Nichol, A. 2021.
\newblock Diffusion models beat gans on image synthesis.
\newblock \emph{Advances in neural information processing systems}, 34: 8780--8794.

\bibitem[{Du and Fan(2018)}]{du2018adaptive}
Du, J.; and Fan, X. 2018.
\newblock Adaptive Watermark Based on Screen.
\newblock In \emph{Proceedings of the 2nd International Conference on Cryptography, Security and Privacy}, 1--4.

\bibitem[{Fang et~al.(2023{\natexlab{a}})Fang, Chen, Qiu, Liu, Xu, Fang, Zhang, and Chang}]{fang2023denol}
Fang, H.; Chen, K.; Qiu, Y.; Liu, J.; Xu, K.; Fang, C.; Zhang, W.; and Chang, E.-C. 2023{\natexlab{a}}.
\newblock DeNoL: A Few-Shot-Sample-Based Decoupling Noise Layer for Cross-channel Watermarking Robustness.
\newblock In \emph{Proceedings of the 31st ACM International Conference on Multimedia}, 7345--7353.

\bibitem[{Fang et~al.(2022)Fang, Jia, Ma, Chang, and Zhang}]{fang2022pimog}
Fang, H.; Jia, Z.; Ma, Z.; Chang, E.-C.; and Zhang, W. 2022.
\newblock Pimog: An effective screen-shooting noise-layer simulation for deep-learning-based watermarking network.
\newblock In \emph{Proceedings of the 30th ACM International Conference on Multimedia}, 2267--2275.

\bibitem[{Fang et~al.(2023{\natexlab{b}})Fang, Qiu, Chen, Zhang, Zhang, and Chang}]{fang2023flow}
Fang, H.; Qiu, Y.; Chen, K.; Zhang, J.; Zhang, W.; and Chang, E.-C. 2023{\natexlab{b}}.
\newblock Flow-based robust watermarking with invertible noise layer for black-box distortions.
\newblock In \emph{Proceedings of the AAAI conference on artificial intelligence}, 5054--5061.

\bibitem[{Fang et~al.(2018)Fang, Zhang, Zhou, Cui, and Yu}]{fang2018screen}
Fang, H.; Zhang, W.; Zhou, H.; Cui, H.; and Yu, N. 2018.
\newblock Screen-shooting resilient watermarking.
\newblock \emph{IEEE Transactions on Information Forensics and Security}, 14(6): 1403--1418.

\bibitem[{Fernandez et~al.(2022)Fernandez, Sablayrolles, Furon, J{\'e}gou, and Douze}]{fernandez2022watermarking}
Fernandez, P.; Sablayrolles, A.; Furon, T.; J{\'e}gou, H.; and Douze, M. 2022.
\newblock Watermarking images in self-supervised latent spaces.
\newblock In \emph{ICASSP 2022-2022 IEEE International Conference on Acoustics, Speech and Signal Processing (ICASSP)}, 3054--3058. IEEE.

\bibitem[{Guan et~al.(2022)Guan, Jing, Deng, Xu, Jiang, Zhang, and Li}]{guan2022deepmih}
Guan, Z.; Jing, J.; Deng, X.; Xu, M.; Jiang, L.; Zhang, Z.; and Li, Y. 2022.
\newblock DeepMIH: Deep invertible network for multiple image hiding.
\newblock \emph{IEEE Transactions on Pattern Analysis and Machine Intelligence}, 45(1): 372--390.

\bibitem[{Guo et~al.(2023)Guo, Zhang, Luo, Guo, Zhang, Su, and Li}]{guo2023practical}
Guo, H.; Zhang, Q.; Luo, J.; Guo, F.; Zhang, W.; Su, X.; and Li, M. 2023.
\newblock Practical Deep Dispersed Watermarking with Synchronization and Fusion.
\newblock In \emph{Proceedings of the 31st ACM International Conference on Multimedia}, 7922--7932.

\bibitem[{Jia et~al.(2020)Jia, Gao, Chen, Hu, Min, Zhai, and Yang}]{jia2020rihoop}
Jia, J.; Gao, Z.; Chen, K.; Hu, M.; Min, X.; Zhai, G.; and Yang, X. 2020.
\newblock RIHOOP: Robust invisible hyperlinks in offline and online photographs.
\newblock \emph{IEEE Transactions on Cybernetics}, 52(7): 7094--7106.

\bibitem[{Jia et~al.(2022)Jia, Gao, Zhu, Min, Zhai, and Yang}]{jia2022learning}
Jia, J.; Gao, Z.; Zhu, D.; Min, X.; Zhai, G.; and Yang, X. 2022.
\newblock Learning invisible markers for hidden codes in offline-to-online photography.
\newblock In \emph{Proceedings of the IEEE/CVF conference on computer vision and pattern recognition}, 2273--2282.

\bibitem[{Jia, Fang, and Zhang(2021)}]{jia2021mbrs}
Jia, Z.; Fang, H.; and Zhang, W. 2021.
\newblock Mbrs: Enhancing robustness of dnn-based watermarking by mini-batch of real and simulated jpeg compression.
\newblock In \emph{Proceedings of the 29th ACM international conference on multimedia}, 41--49.

\bibitem[{Kingma and Ba(2014)}]{kingma2014adam}
Kingma, D.~P.; and Ba, J. 2014.
\newblock Adam: A method for stochastic optimization.
\newblock \emph{arXiv preprint arXiv:1412.6980}.

\bibitem[{Li et~al.(2022)Li, Zhuang, Wang, Liang, Chang, and Yang}]{li2022automated}
Li, C.; Zhuang, B.; Wang, G.; Liang, X.; Chang, X.; and Yang, Y. 2022.
\newblock Automated progressive learning for efficient training of vision transformers.
\newblock In \emph{Proceedings of the IEEE/CVF Conference on Computer Vision and Pattern Recognition}, 12486--12496.

\bibitem[{Li, Liao, and Wu(2024)}]{li2024screen}
Li, Y.; Liao, X.; and Wu, X. 2024.
\newblock Screen-Shooting Resistant Watermarking with Grayscale Deviation Simulation.
\newblock \emph{IEEE Transactions on Multimedia}.

\bibitem[{Liu et~al.(2023)Liu, Si, Qian, Zhang, Li, and Peng}]{liu2023wrap}
Liu, G.; Si, Y.; Qian, Z.; Zhang, X.; Li, S.; and Peng, W. 2023.
\newblock WRAP: Watermarking Approach Robust Against Film-coating upon Printed Photographs.
\newblock In \emph{Proceedings of the 31st ACM International Conference on Multimedia}, 7274--7282.

\bibitem[{Paszke et~al.(2019)Paszke, Gross, Massa, Lerer, Bradbury, Chanan, Killeen, Lin, Gimelshein, Antiga et~al.}]{paszke2019pytorch}
Paszke, A.; Gross, S.; Massa, F.; Lerer, A.; Bradbury, J.; Chanan, G.; Killeen, T.; Lin, Z.; Gimelshein, N.; Antiga, L.; et~al. 2019.
\newblock Pytorch: An imperative style, high-performance deep learning library.
\newblock \emph{Advances in neural information processing systems}, 32.

\bibitem[{Piec and Rauber(2014)}]{piec2014real}
Piec, M.; and Rauber, A. 2014.
\newblock Real-time screen watermarking using overlaying layer.
\newblock In \emph{2014 Ninth International Conference on Availability, Reliability and Security}, 561--570. IEEE.

\bibitem[{Tancik, Mildenhall, and Ng(2020)}]{tancik2020stegastamp}
Tancik, M.; Mildenhall, B.; and Ng, R. 2020.
\newblock Stegastamp: Invisible hyperlinks in physical photographs.
\newblock In \emph{Proceedings of the IEEE/CVF conference on computer vision and pattern recognition}, 2117--2126.

\bibitem[{Tang et~al.(2024)Tang, Wang, Xiang, and Cheung}]{tang2024robust}
Tang, Y.; Wang, C.; Xiang, S.; and Cheung, Y.-m. 2024.
\newblock A Robust Reversible Watermarking Scheme Using Attack-Simulation-Based Adaptive Normalization and Embedding.
\newblock \emph{IEEE Transactions on Information Forensics and Security}.

\bibitem[{Wengrowski and Dana(2019)}]{wengrowski2019light}
Wengrowski, E.; and Dana, K. 2019.
\newblock Light field messaging with deep photographic steganography.
\newblock In \emph{Proceedings of the IEEE/CVF conference on computer vision and pattern recognition}, 1515--1524.

\bibitem[{Xiao et~al.(2024)Xiao, Zhang, Hua, Xia, and Weng}]{xiao2024client}
Xiao, X.; Zhang, Y.; Hua, Z.; Xia, Z.; and Weng, J. 2024.
\newblock Client-Side Embedding of Screen-Shooting Resilient Image Watermarking.
\newblock \emph{IEEE Transactions on Information Forensics and Security}.

\bibitem[{Xu et~al.(2022)Xu, Mou, Hu, Xie, and Zhang}]{xu2022robust}
Xu, Y.; Mou, C.; Hu, Y.; Xie, J.; and Zhang, J. 2022.
\newblock Robust invertible image steganography.
\newblock In \emph{Proceedings of the IEEE/CVF conference on computer vision and pattern recognition}, 7875--7884.

\bibitem[{Zhang et~al.(2018)Zhang, Isola, Efros, Shechtman, and Wang}]{zhang2018unreasonable}
Zhang, R.; Isola, P.; Efros, A.~A.; Shechtman, E.; and Wang, O. 2018.
\newblock The unreasonable effectiveness of deep features as a perceptual metric.
\newblock In \emph{Proceedings of the IEEE conference on computer vision and pattern recognition}, 586--595.

\bibitem[{Zhu et~al.(2018)Zhu, Kaplan, Johnson, and Fei-Fei}]{zhu2018hidden}
Zhu, J.; Kaplan, R.; Johnson, J.; and Fei-Fei, L. 2018.
\newblock Hidden: Hiding data with deep networks.
\newblock In \emph{Proceedings of the European conference on computer vision (ECCV)}, 657--672.

\end{thebibliography}

\end{document}